\documentclass{INTERSPEECH2023}
\usepackage{fixltx2e}
\usepackage{todonotes}
\presetkeys{todonotes}{inline}{}
\usepackage{graphicx}
\usepackage{tipa}
\usepackage{enumitem}
\interspeechcameraready

\newcommand\blfootnote[1]{%
  \begingroup
  \renewcommand\thefootnote{}\footnote{#1}%
  \addtocounter{footnote}{-1}%
  \endgroup
}

\title{DTW-SiameseNet: Dynamic Time Warped Siamese Network for Mispronunciation Detection and Correction}
\name{Raviteja Anantha, Kriti Bhasin$^{\ast}$, Daniela de la Parra Aguilar$^{\ast}$,\\
Prabal Vashisht, Becci Williamson, Srinivas Chappidi}
\address{
Apple
}

\begin{document}

\maketitle
 
\begin{abstract}
% Motivation
Personal Digital Assistants (PDAs) --- such as Siri, Alexa and Google Assistant, to name a few --- play an increasingly important role to access information and complete tasks spanning multiple domains, and by diverse groups of users. A text-to-speech (TTS) module allows PDAs to interact in a natural, human-like manner, and play a vital role when the interaction involves people with visual impairments or other disabilities. To cater to the needs of a diverse set of users, inclusive TTS is important to recognize and pronounce correctly text in different languages and dialects. 
% Problem
% No Citation in InterSpeech Abstract, so cite in Intro section: ~\cite{pss-2014, wavenet-2016}
Despite great progress in speech synthesis, the pronunciation accuracy of named entities in a multi-lingual setting still has a large room for improvement. 
% Shortcomings of existing approaches
Existing approaches to correct named entity (NE) mispronunciations, like retraining Grapheme-to-Phoneme (G2P) models, or maintaining a TTS pronunciation dictionary, require expensive annotation of the ground truth pronunciation, which is also time consuming.
% Approach/Solution/Contributions
In this work, we present a highly-precise, PDA-compatible pronunciation learning framework for the task of TTS mispronunciation detection and correction.
In addition, we also propose a novel mispronunciation detection model called DTW-SiameseNet, which employs metric learning with a Siamese architecture for Dynamic Time Warping (DTW) with triplet loss. 
% Advantages: Locale-agnostic, light-weight and privacy preserving ?
We demonstrate that a locale-agnostic, privacy-preserving solution to the problem of TTS mispronunciation detection is feasible.
% is it possible to quantify? Metrics - Precision, F1 and Pronunciation Accuracy
We evaluate our approach on a real-world dataset, and a corpus of NE pronunciations of an anonymized audio dataset of person names recorded by participants from 10 different locales. Human evaluation shows our proposed approach improves pronunciation accuracy on average by $\approx6\%$ compared to strong phoneme-based and audio-based baselines.
% We evaluate our approach on a real-world dataset, and a corpus of NE pronunciations of an anonymized audio dataset of person names from 10 different locales. Human evaluation shows our proposed approach improves pronunciation accuracy on average by $\approx6\%$ compared to strong phoneme-based and audio-based baselines.
\end{abstract}
\noindent\textbf{Index Terms}: Personal Digital Assistants, Text-to-Speech, Metric Learning, Mispronunciation Detection

% Footnote for equal contributions
\blfootnote{$\ast$ Equal contribution.}

\section{Introduction}
% Set Premise of TTS and it's role in PDAs and user task completion
TTS is an important component in Personal Digital Assistants (PDAs). With the rapid adoption of smart eco-systems and an increase in voice-based applications, PDAs are becoming more common helping users complete tasks. The role of TTS is critical when the interactions involve people with visual impairment or other disabilities. With recent advances in speech synthesis~\cite{wavenet-2016, tacotron-2018, siritts-2021}, current TTS systems can produce expressive and natural sounding voice close to human speech. However, there is significant room for improvement on the multilingual, inclusiveness and personalization aspects. In the digital ecosystems, where it is common to have diverse group of users, multilingual TTS is critical to make users feel acknowledged; and the named entity (NE) pronunciations are particularly important.

% Introduce the problem
In this work we address TTS entity mispronunciations, which can occur because of either:
\begin{itemize}
   \item The NE being a homograph, e.g \textit{David}, which can be pronounced \textipa{/'deI.vid/} for English NEs, or \textipa{/da.'bid/} for Spanish NEs.
   \item The NE has a pronunciation that is difficult to derive from the orthography, and it must still be learned by the TTS system, e.g the Italian name \textit{Palatucci} which is pronounced \textipa{/pa.la.'tu.tSi/}, but can easily be mispronounced by TTS as \textipa{/pa.la.'tuk.si/} if, e.g using a G2P model predominantly trained on Spanish data.
\end{itemize}
TTS personalization can address the former problem, whereas global TTS pronunciation correction is preferable to address the latter. Prior works, which address multilingual and user-specific intonation aspects~\cite{speakerspecifictts-2006, multilingualtts-2020}, require locale-specific models and incur high maintenance cost, especially when working with multiple locales.

We present a locale-agnostic, PDA-compatible, two-stage framework for TTS mispronunciation detection and correction. In the first stage, TTS mispronunciations are detected using a two-step process. First the pronunciation dissimilarity between the user's pronunciation and the TTS pronunciation is computed; second, the dissimilarity score is checked against a threshold to determine if a mispronunciation occurred. The threshold is derived from human labeling to meet the desired precision and recall metrics. In the second stage, the mispronunciation will be qualified for correction (personalization or global learning) using user-engagement signals, such as task completion to ensure precise entity selection, in a privacy-preserving manner. Although we address the problem of TTS mispronunciation, it should be trivial to employ the same framework for correcting ASR NE misrecognitions.

% List contributions clearly at the end of the Intro section
Our contributions can be summarized as:
\begin{itemize}
   \item We propose a highly-precise, locale-agnostic framework for TTS mispronunciation detection and correction by using the correlation between a TTS mispronunciation and the pronunciation dissimilarity of user and TTS pronunciations. 
   \item We present an empirical comparison of phoneme-based algorithms and models along with acoustic models using both intrinsic and extrinsic metrics.
   \item And finally, we introduce a novel mispronunciation detection model called DTW-SiameseNet, which is trained using a metric learning paradigm and learns the distance function via triplet loss to perform Dynamic Time Warping (DTW).
\end{itemize}

\section{Related Work}
Our work is an intersection of three areas: phoneme representation, pronunciation learning, and metric learning.

\subsection{Phoneme Representation}
The task of learning phoneme representations to capture pronunciation similarities is well studied for various downstream applications. A few works have explored the use of phoneme embeddings to perform phonological analogies~\cite{soundanlg-2018}, while others have investigated using embeddings for grapheme-to-phoneme conversion~\cite{jagp-2016}. Improvements in contextual end-to-end Automatic Speech Recognition (ASR) were also realized by using phoneme representations~\cite{ce2easr-2019}.

In this work we apply phoneme embeddings for the task of mispronunciation detection. Recent works show using ASR phoneme embeddings improves mispronunciation detection accuracy~\cite{pemd-2022, pmd-2022}. However, these works use a single phoneme representation (e.g. IPA --- International Phonetic Alphabet), whereas in practice PDAs may use component/task-specific phoneme notation. In our setting, we use two separate phonetic representations, one for ASR and one for TTS. Our goal is to learn dense phoneme representations which capture phonetic similarity within the same phoneme space as well as the relationship between the two different phoneme spaces.

\subsection{Pronunciation Learning}
To learn a correct pronunciation, the first step is to detect a mispronunciation. Prior works~\cite{dnnpl-2014, cnnpl-2019} on mispronunciation detection require a canonical transcription and employ Goodness of Pronunciation (GOP)~\cite{gop-1998}, or classifier based methods.% The detection was mainly used to improve speech recognition. 

A phonological feature-based active-learning method for mispronunciation detection, which estimates phoneme state probabilities using hidden markov models (HMMs) was shown to outperform GOP based methods~\cite{pfmd-2017}, but this still requires locale-specific training and is not feasible for a multilingual setting. A comparison-based approach~\cite{compd-2012} for mispronunciation detection was explored, where two speaker Dynamic Time Warping (DTW) is carried out between student (non-native speaker) and teacher (native speaker) utterances. Unlike this approach where a database of teacher utterances are required and a static distance measure (DTW) is employed, we use a metric-based learning framework, where user and TTS pronunciations are compared using a learned distance function.

% Once a mispronunciation is detected, 

\subsection{Metric Learning}
Metric Learning aims to establish similarity (or dissimilarity) between samples while using an optimal distance metric for learning tasks. Most of the existing metric learning methods rely on learning a Mahalanobis distance~\cite{itml-2007}. The use of a learned distance function in DTW to compare multivariate time series is shown to improve both precision and robustness~\cite{mddtw-2015}. In this work, we adopt a similar strategy to learn a Mahalanobis distance function for audio comparison using DTW. Metric learning uses a linear projection, which limits its ability to learn non-linear characterisitics of the data, so we first apply a non-linear projection using a Siamese architecture and then apply metric learning. To the best of our knowledge, we are the first to use metric learning for mispronunciation detection and correction.

\section{Methods}
\begin{figure*}[t]
 \centering
  \includegraphics[scale=0.32]{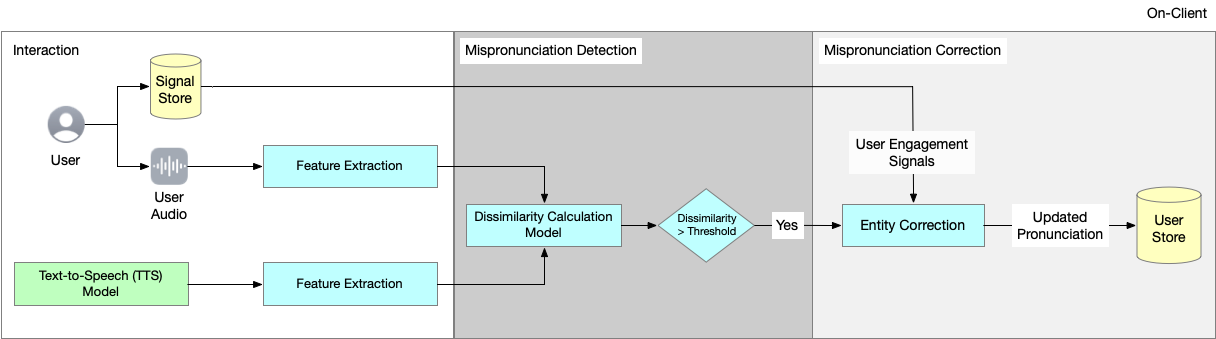}
  \caption{An overview of TTS Mispronunciation Detection and Correction Framework on the Client}
  \label{fig:tmdc-framework}
\vspace*{-0.2in}
\end{figure*}

We introduce a new framework for the task of TTS mispronunciation detection and correction. We propose using the correlation between TTS mispronouncing a NE, and the pronunciation dissimilarity of the user and TTS pronunciations for the same NE exceeding a set threshold. This framework requires us to define a distance function that computes the pronunciation dissimilarity. Once a distance function is obtained, the threshold that correlates with mispronunciation detection with desired precision and recall can be empirically chosen through human labeling. Once a mispronunciation is detected, the TTS entity (e.g. contact name) pronunciation is updated for that specific user, not all users, using the user's pronunciation. An overview of the proposed mispronunciation detection and correction framework is shown in Figure~\ref{fig:tmdc-framework}.

Our experiments for mispronunciation detection can be broadly classified as phoneme-based and audio-based approaches. Pronunciation correction is carried out post mispronunciation detection by using user engagement signals in a privacy-preserving manner. We describe our mispronunciation detection and correction methods below.

\subsection{Phoneme-based Mispronunciation Detection}
\label{section:phoneme-based-methods}
In this section, we elaborate on various methods we evaluated on the TTS mispronunciation detection task where phonemes are used as input.

\subsubsection{Proposed Baseline: P2P Comparison Algorithm}
We present a simple, yet strong baseline called the P2P (Phoneme-to-Phoneme) Comparison algorithm.
In this algorithm, we use:
\begin{itemize}
\item The user interactions on the device to extract the ASR phonemes.
\item The text of the NE as an input to the TTS model to generate the default TTS phonemes.
%\item The mapped ASR phonemes and TTS phonemes to the corresponding XSAMPA phonemes., which is often a one-to-many relationship, in order to see if there is a match between the pronunciations. 
\item The edit distance between the ASR phonemes and TTS phonemes using the Levenshtein distance metric. 
\item Human-labeled data to empirically determine the edit distance threshold based on the desired precision and recall. 
\end{itemize}
% If there is no match between the ASR and TTS pronunciations and the 
If the edit distance is greater than the threshold, the algorithm determines there is a TTS mispronunciation. Once a mispronunciation is detected, we use engagement signals to determine if the pronunciation can be updated with high confidence.
% \todo{Add pseudo code for P2P?}

\subsubsection{Phoneme Embeddings}
\label{section:seq2seq}
% Difference b/w existing works and this work
% Prior works to learn phoneme representations assume a unified phonetic notation standard, typically IPA. However, in practice, PDAs may adopt more than one phonetic notation which can be optimized for a specific task. 

In our setting, ASR and TTS use separate phonesets. As a result, it is not possible to directly compare an ASR phoneme sequence (representation of user's pronunciation) with a TTS phoneme sequence (representation of TTS pronunciation). In addition, these phonesets are locale-specific thereby increasing the number of phonesets.

% Our approach
One simple approach is to use one-hot or multi-hot embeddings, but the resulting representations would be sparse as they do not capture phonetic similarity. Our goals for phoneme embeddings are: (1) obtain dense representations; (2) embeddings should capture phonetic similarity within the same phoneme space, and; (3) capture the relationship between the two different phoneme spaces.

% Question: Add seq2seq+attention diagram ?

% Phoneme Embeddings Model
To accomplish these goals, we train a multi-phoneme sequence-to-sequence (seq2seq) model with multi-head attention~\cite{attention-2017} applied to both the encoder and the decoder. A unidirectional LSTM cell with an output dimension of 100 is used with both the encoder and the decoder. The encoder/decoder attention establishes the corresponding inter-relationship between the input phonemes and the target phonemes, whereas self-attention pays more attention to the intra-relationship of the phoneme pairs in a phoneme sequence.

\subsubsection{GBDT}
We train a Gradient Boosted Decision Tree (GBDT)~\cite{gboosting-2001} classifier using phoneme embeddings as input. For the given user and TTS pronunciations, the phoneme embedding sequences are concatenated and used as input to train a GBDT model using XGBoost~\cite{xgboost-2016} with logistic loss. The annotations are binary labels, where 0 represents both pronunciations are the same, and 1 otherwise. 

\subsubsection{MobileBERT}
We evaluate the MobileBERT~\cite{mbert-2020} architecture, a compressed and optimal version of BERT for resource-limited settings, such as running on mobile devices, with phoneme embeddings as input. MobileBERT is a bidirectional Transformer based on the BERT model. We use the HuggingFace pretrained MobileBERT\footnote{\url{https://huggingface.co/docs/transformers/model_doc/mobilebert}} and conduct knowledge transfer using the multi-head attention from the multi-phoneme seq2seq model described in Section~\ref{section:seq2seq}.

\subsection{Audio-based Mispronunciation Detection}
\label{section:audio-based-methods}
\subsubsection{Dynamic Time Warping}
Dynamic Time Warping (DTW) is an algorithm which can measure the divergence between two time series, in our case audio waveforms, with different phases and lengths. The idea is to compute an optimal warp path between two given waveforms. We use a specific implementation of DTW called FastDTW~\cite{fdtw-2007} as a baseline for audio input.

\subsubsection{Siamese Network using Mel Spectrograms}
Mel-frequency spectrogram is a low-level acoustic representation, which is easily computed from time-domain waveforms. Mel spectrograms are also smoother than raw audio waveforms, which makes them easier to use as features to train a model using variety of loss functions. The sub-waveforms corresponding to entity pronunciation are first extracted using ASR time-spans. We obtain Mel spectrograms of both user and TTS entity pronunciations by applying short-time Fourier transform (STFT) followed by a nonlinear transform to the frequency axis of the STFT. This representation using Mel frequency scale emphasizes details in lower frequencies, which are critical to speech intelligibility. 

We use a Siamese neural network~\cite{siamese-1993}, which consists of twin networks and the parameters between them are tied. We use convolutional layers in twin networks which accept two Mel spectrograms as inputs and determines whether they are similar. We use 3 channels with filters of varying size and fixed stride length of 1. We use ReLU for activation function and max pooling. The outputs of convolutional layers are flattened and concatenated before passing on to a sigmoid activation function. We use the Adam optimizer and cross-entropy loss to learn binary classification.

\subsubsection{Proposed Method: DTW-SiameseNet}
We propose a novel mispronunciation detection model that employs metric learning with a Siamese architecture for DTW with a triplet loss. We use Mahalanobis distance as our metric; non-Mahalanobis based metric learning was also proposed but this suffered from non-convexity or computational complexity~\cite{itml-2007}. Given two \textit{d}-dimensional vectors \textit{x} and \textit{y}, the square Mahalanobis distance parametrized by a symmetric Positive Definite (PD) matrix \textit{A} between the two vectors is defined as:
\begin{equation} \label{eq:1}
D_{A}(x, y) = (x - y)^{T} A (x - y).
\end{equation}
The Positive Definiteness of the matrix \textit{A} guarantees that the distance function will return a positive distance. The Mahalanobis matrix \textit{A} can be decomposed as:
\begin{equation} \label{eq:2}
A = G^{T}G.
\end{equation}
This can be interpreted as \textit{G} being distributed to (\textit{x} - \textit{y}) terms, i.e., linear transformation applied to the input. Our goal is to learn the PD matrix \textit{A} based on some constraints over the distance function. We apply two constraints:

\begin{itemize}
   \item If two vectors are similar then the distance metric D(.) is smaller than an upper bound u\textsubscript{bound}, and;
   \item If two vectors are dissimilar then the distance metric D(.) is greater than a lower bound l\textsubscript{bound}.
\end{itemize}

We combine the two constraints into a triplet constraint. Given three \textit{d}-dimensional vectors \textit{x}, \textit{y} and \textit{z}; where \textit{x}, \textit{y} are similar and \textit{x}, \textit{z} are dissimilar we express the constraint as:

\begin{equation} \label{eq:3}
D_{A}(x, y) - D_{A}(x, z) < -\rho,
\end{equation}
where $ 0 < \rho < l_{bound} - u_{bound} $.

We apply non-linear projection using the Siamese architecture on the inputs before linear projection through \textit{A}. We use a unidirectional LSTM with an attention mechanism, $f_{W}(.)$, for the twin networks, where the parameters \textit{W} are tied. For given inputs \textit{x}, \textit{y} with a randomly drawn \textit{z}, the objective function is defined as

\begin{equation} \label{eq:4}
l(A, W) = \rho + D_{A}(f_{W}(x), f_{W}(y)) - D_{A}(f_{W}(x), f_{W}(z)).
\end{equation}

The overall loss is given as:
\begin{equation} \label{eq:5}
L(A, W) = \sum_{t} l(A, W).
\end{equation}

We use SGD to update the parameters \textit{W} and learn the Mahalanobis matrix \textit{A}, which together constitute the distance function. Since we need the updates to \textit{A} to be gradual and stable, we add a regularization term. LogDet divergence~\cite{logdet-2006} is shown to be the most optimal for regularizing the metric learning process and is invariant to linear group transformations. The LogDet divergence for \textit{A} and $A_{t}$ (\textit{A} at time-step \textit{t}) is given as:
\begin{equation} \label{eq:6}
D_{ld} (A, A_{t}) = tr(A A_{t}^{-1}) - \log (\det (A A_{t}^{-1}) ) - d.
\end{equation}

Applying the LogDet divergence the metric learning model for updating \textit{A} will be 
\begin{equation} \label{eq:7}
A_{t+1} = \arg \min_{A \succ 0}  D_{ld} (A, A_{t}) + \eta_{t} l(A, W),
\end{equation}
where $\eta_{t} > 0$ is a regularization parameter that balances LogDet regularization function $D_{ld} (A, A_{t})$.

Once the distance function is learned, we use it to compute the distance between the inputs using the traditional DTW algorithm as shown below, where we use a moving window of dimension $d$ to choose input sub-sequences:

\begin{equation}  \label{eq:8}
  D_{A}(i,j) = D_{A}(x^{i}, y^{i}) + min
    \begin{cases}
      D_{A}(i-1, j-1) \\
      D_{A}(i-1, j) \\
      D_{A}(i, j-1).
    \end{cases}       
\end{equation}

The main difference between the traditional DTW algorithm and DTW-SiameseNet lies in the fact that we learn the distance function \textit{D}(.) parameterized on \textit{A} and \textit{W} for the inputs \textit{x} and \textit{y}.

\subsection{Pronunciation Correction}
Once the pronunciation dissimilarity score is computed, and if it meets the chosen threshold, we deem the TTS pronunciation as a mispronunciation. We employ user engagement signals, such as task completion, to avoid incorrectly updating the pronunciation of an entity. For example, if the task was to call a person, prior to updating the contact pronunciation, we check if the call was successful and the call duration was greater than a predetermined number of seconds. 

\section{Training Data}
\label{section:data}
We use two datasets: one real-world (phoneme-based) and one human-generated (audio) NE pronunciations dataset; comprised of data from 10 locales to train and evaluate phoneme-based and audio-based methods.

\subsection{Phoneme-based Dataset}
\label{section:phoneme-based-data}
We curated a real-world dataset comprised of 50K randomized and anonymized user requests from 10 different locales, where each request contain a reference to an entity. This dataset is used to train and test phoneme-based approaches described in Section~\ref{section:phoneme-based-methods}. Each locale has 5K data points with ASR and TTS phoneme representations for the entity pronunciation, but no user audio. On average, 30\% of entity names in each locale are non-native names and $>$20\% are homographs. This dataset has mispronunciations in the range of 15\% to 28\%.  

\subsection{Audio Dataset}
\label{section:audio-data}
We created an anonymized audio dataset comprised of 30K audio requests using human annotators. Each locale has 1K unique entities with person, location and business names. Human participants are provided with prompts, such as ``Directions to X" or ``Call X", which are used to record the audio. Each entity gets audio from 3 different participants to capture variance from different genders and age groups. Since we did not use locale-specific participants, this dataset contains 40\% to 50\% of human mispronunciations. On average, ~22\% of the names are homographs with ~17\% being non-native names. 

\section{Results}
%Add stat-sig results, both en and non-en locales}
Below we present both intrinsic and extrinsic metrics, unless specified otherwise metrics for methods described in Section~\ref{section:phoneme-based-methods} and \ref{section:audio-based-methods} are computed using data described in Section~\ref{section:phoneme-based-data} and \ref{section:audio-data} respectively.
%\subsection{Intrinsic Metrics}
\begin{table}[th]
  \caption{Intrinsic Metrics average across the 10 locales: en-US, en-CA, en-GB, en-AU, en-IN, fr-FR, es-ES, es-MX, es-US, ja-JP. All the differences among methods are statistically significant.}
  \resizebox{\columnwidth}{!}{
  \label{tab:intrinsic-metrics}
  \centering
  \begin{tabular}{l l  c  c }
    \toprule
    \multicolumn{1}{l}{\textbf{Data Type}} & \multicolumn{1}{l}{\textbf{Method}} & 
                                         \multicolumn{1}{c}{\textbf{Precision}} & 
                                         \multicolumn{1}{c}{\textbf{Recall}} \\
    \midrule
    &  P2P               & $95.29 (\pm0.01)$ & $72.87 (\pm0.01)$       \\
    Phoneme-based & GBDT              & $\textbf{95.78}(\pm0.04)$  & $\textbf{94.91}(\pm0.02)$    \\
    &  MobileBERT        & $94.22 (\pm0.15)$  & $92.36(\pm0.19)$      \\
    \midrule
    &  DTW               & $62.5 (\pm0.01)$  & $30.64 (\pm0.01)$       \\
    Audio-based &  SiameseNet        & $94.77 (\pm0.27)$  & $91.12 (\pm0.18)$      \\
    &  DTW-SiameseNet    & $\textbf{95.17}(\pm0.12)$  & $\textbf{93.87}(\pm0.08)$    \\
    \bottomrule
  \end{tabular}
  }
\end{table}

%\subsection{Extrinsic Metrics}
We compute pronunciation accuracy using both percentage and a 3-point Likert scale, where in the latter 1 represents correct entity and TTS pronunciations are different, 2 represents a partial similarity, and 3 represents full similarity. We use a TTS system with an average pronunciation accuracy of 88\%. 

\begin{table}[th]
  \caption{Pronunciation accuracy (extrinsic metric) across the 10 locales on audio-based dataset using DTW-SiameseNet.}
  \resizebox{\columnwidth}{!}{
  \label{tab:intrinsic-metrics}
  \centering
  \begin{tabular}{l c c c c }
    \toprule
    \multicolumn{1}{c}{\textbf{Scale}} & \multicolumn{1}{c}{\textbf{en-US/CA/GB/AU}} & \multicolumn{1}{c}{\textbf{fr-FR}} & \multicolumn{1}{c}{\textbf{es-ES/MX/US}} & \multicolumn{1}{c}{\textbf{ja-JP}}\\
    \midrule
    Percent & $94.34$ & $92.89$ & $93.25$ & $90.17$\\
    Likert (1-3) & $2.83$ & $2.79$ & $2.80$ & $2.61$\\
    \bottomrule
   \end{tabular}
  }
\end{table}

\section{Conclusion}
% - Further room for filtering out Stutter from input audio and reduce dependency on phonemes
% - Cross-lingual pronunciation (audio) synthesis from User audio
In this paper, we presented a locale-agnostic framework for TTS mispronunciation detection and correction, which is compatible with PDAs. In addition, we described a novel metric learning model for audio comparison called DTW-SiameseNet. We investigated and presented empirical comparison of various phoneme and audio based methods. 

% \section{Acknowledgements}
% This section is intentionally left blank

\bibliographystyle{IEEEtran}
\bibliography{mybib}

\end{document}